\documentclass[10pt, a4paper]{article}

\usepackage[final]{lrec2026} 

\usepackage{amsmath}

\title{Human Label Variation in Implicit Discourse Relation Recognition}

\name{Frances Yung,$^{1,*}$ Daniil Ignatev,$^{2,*}$ Merel Scholman,$^{2}$ \\
{\large \textbf{Vera Demberg},$^{1}$}
{\large \textbf{Massimo Poesio}$^{2,3}$}} 

\address{
  $^{1}$Saarland University, Saarland Informatics Campus, Germany\\
  $^{2}$Utrecht University, Utrecht, The Netherlands\\
  $^{3}$Queen Mary University of London, London, UK\\
  \{frances,vera\}@coli.uni-saarland.de \\
  \{d.ignatev,m.c.j.scholman,m.poesio\}@uu.nl \\
  \vspace{0.2cm}
  $^*$ Contributed equally to this submission.}

\abstract{There is growing recognition that many NLP tasks lack a single ground truth, as human judgments reflect diverse perspectives. To capture this variation, models have been developed to predict full annotation distributions rather than majority labels, while perspectivist models aim to reproduce the interpretations of individual annotators. In this work, we compare these approaches on Implicit Discourse Relation Recognition (IDRR), a highly ambiguous task where disagreement often arises from cognitive complexity rather than ideological bias. Our experiments show that existing annotator-specific models perform poorly in IDRR unless ambiguity is reduced, whereas models trained on label distributions yield more stable predictions. Further analysis indicates that frequent cognitively demanding cases drive inconsistency in human interpretation, posing challenges for perspectivist modeling in IDRR.
\\ \newline  }

\begin{document}

\maketitleabstract

\section{Introduction}

Discourse annotation is a complex task that is prone to significant disagreement even among trained annotators \cite{artstein-poesio-2008-survey,SpoorenDegand+2010+241+266}. This is particularly true for the annotation of implicit discourse relations (DRs) 
which are not marked by explicit discourse connectives, such as \textit{because} for \textit{reason} or \textit{whereas} for \textit{contrast} \cite{prasad2017penn,hoek-etal-2021-less}.

\newcounter{excounter}
\newcommand{\ex}[1]{%
  \refstepcounter{excounter}%
  \begin{list}{}{\leftmargin=1.5em \rightmargin=0pt}%
  \item[(\theexcounter)] {\small #1}%
  \end{list}%
}

\ex {Arg1: \textit{Our planet will not have enough room for two faction, both of which strive to dominate the whole world.} Arg2: One of them \underline{must} give way.}

Annotating DRs involves high-level understanding of textual semantics and pragmatics in natural language.  Disagreements arise from various factors, including linguistic ambiguity. For instance, the analysis of Example (1) may vary depending on whether \textit{"must"} implies an \textit{obligation} (prioritizing the evaluative aspect) or \textit{inevitability} (prioritizing causality). This allows for multiple readings of the same passage and suggests that adopting a single ground-truth reading would lead to information loss.

Nonetheless, implicit discourse relation recognition (IDRR) has traditionally been a single-label classification task in NLP.  The largest discourse annotated resource, the Penn Discourse Treebank \citelanguageresource[PDTB,][]{AB2/SUU9CB_2019}, is primarily annotated with one label per sample and includes alternative labels in less than $5\%$ of the samples, even though disagreement between two experts is expected to be around $44\%$ \cite{hoek-etal-2021-less}. In recent years, the performance of novel IDRR systems appears to have plateaued \cite[SOTA on PDTB-3: F1 $55\%$ ACC $64\%$][]{wang-etal-2025-using}. This may indicate that a more flexible approach that permits concurrent interpretations is necessary to ensure a more complete solution. Several studies have reported the benefits of training from multiple annotations in IDRR \cite{yung-etal-2022-label,costa-kosseim-2024-exploring,long-etal-2024-multi}.

Disagreement-aware approaches in data collection and machine learning have earned recognition in various NLP tasks \cite{basile2020s,uma2021learning,rottger-etal-2022-two,Cabitza_Campagner_Basile_2023,xu2026beyond},
particularly in tasks involving subjectivity and linguistic ambiguity. 
Specialized architectures have been proposed to model individual annotator stances instead of the ground truth in such domains as hate speech detection and toxicity detection \cite{davani-etal-2022-dealing,kanclerz-etal-2022-ground,ngo-etal-2022-studemo,deng-etal-2023-annotate,mokhberian-etal-2024-capturing}. 
This approach could tentatively apply to crowd-sourced discourse annotations, as labelers can interpret the pragmatic function of a text subjectively; however, it may entail problematic implications, as human-produced data is known to be prone to errors or deliberate attempts at brute-forcing the task by annotators \cite{poesio&artstein:ACL-ANNO-05,plank2014linguistically,plank-2022-problem}. The applicability of perspectivism is therefore task-dependent. 
In particular, since variation in DR interpretation often stems from individual differences in cognitive ability \cite{sanders-etal-2022-towards, SpoorenDegand+2010+241+266}, the applicability of existing perspectivist models tailored for subjective tasks remains an open question.

This study examines the performance of various IDRR classifiers for English with different levels of awareness of label disagreement, including classical approaches that are optimized to predict the most probable label, models that are trained to predict label distributions, and perspectivist models that make annotator-specific prediction.  We make use of the DiscoGeM dataset \citetlanguageresource{scholman-etal-2022-discogem,yung-demberg-2025-crowdsourcing}, a large collection of unaggregated crowdsourced labels for English implicit DRs. While training on label distributions has been shown to improve single-label prediction performance \cite{yung-etal-2022-label, pyatkin-etal-2023-design,costa-kosseim-2024-exploring}, the robustness and generalizability of these models across different tasks remain unclear. For instance, it is not yet known whether the probability outputs of a single-label prediction model naturally align with label-distribution prediction, or whether annotator-specific prediction models can effectively learn IDRR given a limited number of annotator-specific training instances. Our experiments aim to address these open questions through systematic cross-task evaluation. Our contributions include:
\begin{enumerate}
    \item  We confirmed that disagreement-aware approaches improve the accuracy of majority-label prediction.
    \item We find that the effectiveness of annotator-specific models strongly depends on label granularity ($5$ vs. $17$ classes).
     \item We demonstrate that learning from label distributions provides the most reliable performance for predicting fine-grained label distributions, when the specific annotators are unknown.
    \item Our analysis reveals that annotator inconsistency in cognitively demanding cases, which are common in IDRR, substantially reduces model performance, particularly in ambiguous fine-grained settings.
\end{enumerate}




\section{Related work}

\paragraph{Disagreement and perspectives in annotation} 
NLP tasks, such as sentiment analysis, toxicity detection, and hate speech detection, 
involve subjective evaluation of a parameter by annotators. This subjectivity challenges the existence of a ground truth (\textit{alias} gold standard) in the respective datasets. Divergent annotations may need to be treated as a meaningful signal rather than noise \citep{aroyo&welty:AIMagazine15,pavlick2019inherent, larimore-etal-2021-reconsidering, plank-2022-problem}. Specifically, these annotations can reflect the multitude of the labelers' backgrounds and therefore offer a more representative image of subjective data compared to the aggregated gold standard \citep{diaz2018addressing, garten2019incorporating, akhtar2019new, ferracane-etal-2021-answer}. Several studies delved into the reasons behind disagreement in detail, identifying various pattern dependent on task \cite{nie-etal-2020-learn,jiang-marneffe-2022-investigating,jiang-etal-2023-ecologically,sandri-etal-2023-dont}.
%
For example, annotators' labeling of toxicity and hate speech vary notably depending on their identity, age, personality-related variables, beliefs and stereotypes \cite[][ see also \citet{frenda2024perspectivist} for an overview]{sap2022annotators,sang2022origin,davani2023hate}.

\paragraph {Annotator-specific label prediction models}
To embrace the distinct points of view on subjective tasks and model perspectives in annotation, researchers call for the preservation of diverging labels in data collection  \citep{aroyo&welty:AIMagazine15,poesio-et-al:NAACL19,basile2020s,prabhakaran2021releasing,Cabitza_Campagner_Basile_2023}.
%
Existing perspectivist models enhance transformer-based classifiers with trainable annotator-specific components, such as classification heads \cite{davani-etal-2022-dealing}, or embeddings \cite{deng-etal-2023-annotate}. These architectures enable the integration of annotator-level features, including basic embedded annotator identifiers, and also relevant sociodemographic variables \cite{mokhberian-etal-2024-capturing}. We will further explain the models used in our experiments in Section~\ref{sec:models}.

Nonetheless, human label variation does not solely stem from valid differences in perspective but can also result from annotation errors caused by misinterpretation. \citet{weber2024-varierrnli} propose an additional round of explanation annotation and subsequent analysis to tease apart annotation errors and genuine perspectivist label variations.  
In this work, we conduct a manual analysis to estimate the proportion of annotation variation attributable to differences in interpretation versus annotation errors in IDRR, using third-person annotations of the cues underlying each assigned label.


\paragraph{Perspectives in IDRR} 
Low agreement in implicit DR annotation has been widely reported in existing work \cite{das-etal-2017-good,hoek-etal-2021-less}, also \citelanguageresource{polakova2013introducing}.  Unlike in other subjective NLP tasks, variation in DR interpretation has not been linked to annotators' demographic factors. Instead, disagreement has been shown to arise from relation-specific ambiguities, such as whether the relation contains a 1st/2nd person conceptualizer 
\cite{sanders1992toward,SpoorenDegand+2010+241+266}. 
Additional sources of disagreement include 
textual ambiguity,
differences in background knowledge,
and ambiguities in relation definitions
\cite{zikanova2024text,zikanova2025gold,hewett-stede-2025-disagreements,ignatev2025annotator}. Several existing resources provide consistent multiple annotations designed to study disagreement in discourse interpretation \citelanguageresource{hewett-stede-2025-disagreements,zeldes-etal-2025-erst,zikanova2025gold}. In this work, we use DiscoGeM, a dataset focused on \textbf{implicit} DRs.


Recent studies have explored learning from label distributions
\cite{yung-etal-2022-label,pyatkin-etal-2023-design,costa-kosseim-2024-exploring,long-etal-2024-multi}, and have consistently shown advantages over majority-label training. However, perspectivist models that explicitly predict the annotations of individual workers have not yet been applied to IDRR, which is a gap this work seeks to address.



\section{Experiments}

The objective of our experiments is to evaluate how well models can predict disagreement in English IDRR at different levels of granularity. Specifically, we consider three tasks: \textbf{single-label prediction}, \textbf{label-distribution prediction}, and \textbf{annotator-specific label prediction}. We systematically compare models optimized for each of these tasks and evaluate their performance across all three tasks.

\subsection{Data}
\label{sec:dataset}
To train IDRR models that account for label variation, we use the DiscoGeM~1.5 dataset \citetlanguageresource{scholman-etal-2022-discogem,yung-demberg-2025-crowdsourcing}, which is the only available resource of PDTB-style \textbf{implicit} DRs that provides multiple annotations for each instance.
DiscoGeM is an English multi-genre corpus, containing excerpts from European Parliament proceedings, Wikipedia, and novels. 
Labels were collected through crowd-sourcing, in which annotators with no presumed knowledge of DRs 
select a connective that fits between sentences 1 and 2 (Arg1 and Arg2) from a predefined list.
The workers can opt for the "\textit{(no direct relation)}" label if they cannot identify any relation between the arguments.


DiscoGeM follows the sense classification of the PDTB, which is structured as a three-level hierarchy, with four coarse-grained Level-1 sense categories and more fine-grained senses for each of the subsequent levels \citep{webber2019pdtb3}. The annotation interface allows for annotation to the finest-grained Level-3 senses ($29$ classes), each represented by a connective choice.
On the other hand, our experiments adopt the widespread practice of Level-1 ($5$ classes) and Level-2 ($17$ classes) DR prediction \cite{xiang2023survey}.

Table~\ref{tab:datasets} compares the statistics of DiscoGeM with other multi-annotator corpora previously used for perspectivist modeling. 
We observe that the mean number of items per worker in DiscoGeM is comparable to- and even surpasses MDA. At the same time, the complexity of the task can make it hard to learn the workers' individual stances even from that many examples.
In particular, at Level-3 granularity, the average count of distinct labels per instance amounts to $6.3$, leading to a relatively low agreement rate of $0.404$ and high entropy of $0.86$.\footnote{Considering the $10$ collected labels per instance but not the complete set of $29$ Level-3 labels, for which most have $0$ votes.} 
Such high level of label variation is not observed in other disagreement datasets.


\begin{table}[t]
\small
\centering
\begin{tabular}{l|cccc}
\hline
Col & DG1.5 & MDA & HSB & GHC \\
\hline
Train & 4.5/52k & 6.59/33k & 0.78/4.7k & 22/69k \\
Dev & 0.6/7k & - & - & - \\
Test & 1.2/14k & 3/11k & 0.16/1k & 5.5/17k \\
\#W & 164 & 334 & 6 & 18 \\
\#A/\#W & 448 & 160 & 1120 & 4807 \\
\#A/\#I & 10 & 5 & 6 & 3.13 \\
\#L & 17/5 & 2 & 2 & 2 \\
\hline
\end{tabular}
\caption{Statistics of DiscoGeM 1.5 compared to Multi-Domain Agreement, MDA \cite{leonardelli-etal-2021-agreeing}; Hate Speech-Brexit, HSB \cite{akhtar2021whose}; and GHC \cite{kennedy2022introducing}; counts reported as items/annotations. \#W: worker count; \#A/\#W: average annotation/worker ratio; \#A/\#I: average annotation/instance; \#L:  unique label count.}
\label{tab:datasets}
\end{table}


\subsection{Models}
\label{sec:models}

We compare three classes of models across the three IDRR tasks.
To ensure consistency and to facilitate comparison with the existing IDRR pipelines \cite{long-etal-2024-multi,costa-kosseim-2024-exploring}, we use RoBERTa-base \cite{liu2019roberta} as a backbone for all architectures. Separate models are trained for Level-1 and Level-2 DR classification. 

\noindent \paragraph{Single truth (ST) model} Our baseline is a RoBERTa-base classifier trained on the majority label of each instance. We derive the model’s normalized probability distribution (\textbf{ST.logit}) to represent its single-label prediction for the label-distribution task. For annotator-specific label prediction, the model’s predicted single label (\textbf{ST.top1}) is considered as the model's prediction of every perspective for the given instance.

\noindent \paragraph{Soft label models}
We experiment with two models that allow soft-label prediction. The first method performs multi-label classification in the form of a binary classification for each label type. Specifically, we use a Binary Cross-Entropy (BCE) loss averaged over batch and labels, which is a standard approach for multi-label text classification \cite[e.g.,][]{he2018joint,huang-etal-2021-balancing}. Following \citet{pyatkin-etal-2023-design}, labels with 20\% or more distribution in each item of DiscoGeM are selected as the multiple gold labels during training. Similarly to the ST model, we use probability distributions (\textbf{multi.logit}) to represent predictions for the label-distribution task.

 The other soft prediction model we explored directly predicts the full annotation distribution (\textbf{label-dist}) of each sample based on a soft loss, specifically the Kullback-Leibler divergence loss, following \citet{fornaciari2021beyond, uma2021learning}. 
 For single label and annotator-specific label prediction tasks, labels with the highest probability (\textbf{multi.top1} and \textbf{dist.top1}) are used as the single-label prediction.

\paragraph{Perspectivist models}
We compared two perspectivist models in our experiments: the multi-task-based multi-annotator model \cite[\textbf{MT},][]{davani-etal-2022-dealing} and the Annotator Embeddings model \cite[\textbf{AE},][]{deng-etal-2023-annotate}. Both are trained on paired annotator IDs and their specific labels provided in DiscoGeM and are applied to predict the annotations of specific annotators in the test set.


The \textbf{MT} model learns multiple perspectives as separate tasks, using a separate head for each annotator in the dataset on top of a shared encoder. During training, the model is jointly optimized against all labels assigned to a single instance, whereby Cross-Entropy is computed separately for each head and then summed. 
The \textbf{AE} model models worker-specific annotation patterns using distinct annotator and annotation embeddings, which are integrated with the original text embeddings through a weighted sum. The statistics of the original data used in the MT and AE models are shown in Table~\ref{tab:datasets}.



We additionally compare against majority-vote baselines (\textbf{MT.maj} and \textbf{AE.maj}), which assign the most frequently predicted label per item to all annotators. These outputs are also used in the single-label prediction setting. For the label-distribution prediction task, we average the predicted probability distributions of the ten annotators in each test sample, producing one aggregated label distribution per item (\textbf{MT.logit} and \textbf{AE.logit}).
\section{Results}

\paragraph{Single-label prediction}
Table~\ref{tab:single_pred} presents the results for the single-label prediction task, evaluated using standard macro-F1 and accuracy against the single majority gold label provided in DiscoGeM. The results replicate previous findings claiming that training on label distributions can improve single-label prediction performance \cite{yung-etal-2022-label,pyatkin-etal-2023-design,costa-kosseim-2024-exploring}. The relatively lower performance of the multi-label prediction model on Level-2 DR prediction may be attributed to the increased randomness involved in selecting the top label among a larger set of predicted labels. In contrast, the perspectivist AE model further outperforms the soft-label approaches on Level-1 prediction, but this advantage does not extend to Level-2, suggesting that a higher number of classes may constrain model performance.

\begin{table}[htpb]
    \centering \small
    \begin{tabular}{@{}l|llll@{}}
        \hline
        & F1$_{Lv1}$ & Acc$_{Lv1}$ & F1$_{Lv2}$ & Acc$_{Lv2}$ \\
        \hline
        ST.top1 & 0.48 & 0.58 & 0.24 & 0.44 \\
        multi.top1 & 0.47 & 0.57 & 0.21 & 0.29 \\
        dist.top1 & 0.49 & 0.6 & \textbf{0.28}* & \textbf{0.47} \\
        MT.maj \ & 0.36 & 0.39 & 0.03 & 0.03 \\
        AE.maj & \textbf{0.56}* & \textbf{0.7}* & 0.21 & 0.4 \\
        \hline

    \end{tabular}
    \caption{\small Results of the \textbf{single label prediction task} evaluated by accuracy and macro-F1. Performance averaged per 3 runs. The highest score shown in bold. * marks significant advantage over ST, $p < 0.05$. Here and further on, significance is measured with a resampled paired t-test \cite{10.1162/089976698300017197}.
    }
    \label{tab:single_pred}
\end{table}

\paragraph{Label-distribution prediction}
\begin{table}[htpb]
    \centering \small
    \begin{tabular}{l|llll}
         \hline
        Lv1 & CE & JSD & MD & ED \\
        \hline
        ST.logit & 1.708 & 0.748 & 0.570 & 0.566 \\
        multi.logit & 1.520* & 0.153* & 0.630 & 0.356* \\
        label-dist & \textbf{1.341}* & \textbf{0.118}* & 0.547* & 0.304* \\
        AE.logit & 1.402* & 0.123* & \textbf{0.367}* & \textbf{0.216}* \\
        \hline

         Lv2 & CE & JSD & MD & ED \\
        \hline
        ST.logit & 3.071&0.371& 1.126& 0.551 \\
        multi.logit &2.788& 0.342& 1.021* & 0.440* \\
        label-dist &\textbf{2.265}* & \textbf{0.269}* & \textbf{0.878}* & \textbf{0.347}* \\
        AE.logit & 2.947& 0.450& 1.187& 0.488\\
        \hline
        
    \end{tabular}
    \caption{\small Results of the \textbf{label distribution prediction task} evaluated by $4$ soft metrics. Results based on $3$ random seeds. The lowest score shown in bold. * marks significant advantage over ST, $p < 0.05$.}
    \label{tab:dist_results}
\end{table}

Table~\ref{tab:dist_results} compares the results where the predicted distributions are evaluated against the gold annotation distributions of each instance using four soft metrics: Cross-Entropy (CE), Jensen–Shannon Divergence (JSD), Manhattan Distance (MD), and Euclidean Distance (ED). Lower values indicate greater similarity between the predicted and reference distributions.
The results show that the \textbf{ST.logit} model lags behind the others, suggesting that the probability distribution of the single-truth model does not effectively capture the underlying label distribution, and that training with soft labels is beneficial. The \textbf{label-dist} model achieves the best overall performance in this task, which is expected given that it is explicitly optimized for distributional prediction. The perspectivist models perform comparably well, but only at Level-1, again indicating that their advantage diminishes as the number of classes increases.

\paragraph{Annotator-specific label prediction}
\begin{table}[t]
    \centering \small
    \begin{tabular}{@{}l|llll@{}}
        \hline
        & F1$_{Lv1}$ & Acc$_{Lv1}$ & F1$_{Lv2}$ & Acc$_{Lv2}$ \\
        \hline
        ST.top1 & 0.35 & 0.44 & 0.15 & 0.28 \\
        multi.top1 & 0.35 & 0.44 & 0.14 & 0.20 \\
        dist.top1 & 0.35 & 0.46* & 0.17* & 0.29* \\
        MT \ & 0.33 & 0.34 & 0.08 & 0.10 \\
        AE.maj & 0.37* & 0.50* & 0.11 & 0.21 \\
        AE & \textbf{0.75}* & \textbf{0.84}* & \textbf{0.23}* & \textbf{0.44}* \\
        \hline
    \end{tabular}
    \caption{\small Results of the \textbf{annotator-specific label prediction task} evaluated by accuracy and macro-F1, considering each annotator-specific annotation as one instance. Performance averaged per 3 runs. The highest score shown in bold. * marks significant advantage over ST, $p < 0.05$.
    }
    \label{tab:hard}
\end{table}

Finally, Table~\ref{tab:hard} presents the results of annotator-specific label prediction, averaged over 3 runs. The models are evaluated using \textit{global} accuracy and macro F1-score, where each annotator-specific annotation is counted as an individual instance \cite{deng-etal-2023-annotate}.
The  results suggest that the perspectivist \textbf{AE} model consistently outperforms all other models by a substantial margin. Although the performance on Level-2 prediction is substantially lower than on Level-1, as also observed in the previous tasks, it still remains approximately twice as high as that of the other models. The low performance of \textbf{AE.maj} highlights the disagreement between the annotators and demonstrates the importance of annotator-specific prediction. 
The \textbf{MT} model, in contrast, appears unable to predict annotation perspectives in this task. Notably, the original model was tested on 2-way and 6-way classification with fewer annotations per instance ($3$ vs $10$), and high number and diversity of labels could have exacerbated the weaknesses of multi-task learning, such as conflicting gradient updates \cite{yu2020gradient}. 


To summarize, the results show that:
(1) learning from label distributions or annotator-specific labels improves performance in single-label prediction, whereas single-label strategies perform poorly on distributional prediction tasks due to the high disagreement and ambiguity in IDRR;
(2) models optimized for soft-label prediction (i.e., to learn per-sample label distributions without annotator information) and perspectivist models that are optimized for annotator-specific predictions each achieve the best performance on their respective tasks; and
(3) perspectivist models can also successfully  predict label distributions, but their effectiveness decreases as the number of classes increases.

\section{Analysis}
The AE model outperforms the label-dist model only on Level-1 in distribution predction (Table~\ref{tab:dist_results}); its performance on annotator-specific prediction also drops more significantly from Level-1 to Level-2 prediction compared with other models (Table~\ref{tab:hard}). These results 
show that the perspectivist model struggles to predict fine-grained annotation of specific annotators.
In this section, we aim to identify the conditions under which the perspective-aware models can accurately predict a worker’s individual label.
In addition, we will analyze the different types of perspectives and biases in IDRR.

\subsection{When do perspectivist models perform better?}

During perspective-aware training, models can be biased in favor of more prolific annotators.
To assess the effect of this factor, 
we divide the annotators into two groups based on the counts of their items in the train set: less or equal to the median count (AE$_{\#E-}$, mean count=93) or above the median (AE$_{\#E+}$, mean count=540). The model performance on these subsets of workers is shown in the upper half of Table~\ref{tab:ablation}. Surprisingly, comparing the metrics across the two groups reveals no significant difference, suggesting that the per-annotator training samples are enough in the dataset and the difference in model performance is not due to the prolificness of the workers.

%

\begin{table}[t]
    \centering \small
    \begin{tabular}{@{}l|llll@{}}
        \hline
        & F1$_{Lv1}$ & Acc$_{Lv1}$ & F1$_{Lv2}$ & Acc$_{Lv2}$ \\
        \hline
        AE$_{\#E+}$ & 0.75 & 0.84 & \textbf{0.23} & \textbf{0.44} \\
        AE$_{\#E-}$ & \textbf{0.76} & \textbf{0.85} & \textbf{0.23} & 0.43 \\
        \hline
        AE$_{Low-\kappa}$ & 0.68 & 0.81 & 0.14 & 0.25 \\ 
        AE$_{Med.-\kappa}$  & 0.74 & 0.83 & 0.22 & \textbf{0.46} \\ 
        AE$_{High-\kappa}$ & \textbf{0.76} & \textbf{0.85} & \textbf{0.24} & 0.44 \\ 
        \hline
    \end{tabular}
    \caption{\small Results of the \textbf{worker-specific label prediction task} in different subsets of workers. Performance averaged per 10 runs. The highest scores shown in bold. 
    }
    \label{tab:ablation}
\end{table}

\begin{figure}[htpb]
    \centering \small
    \includegraphics[width=0.9\linewidth]{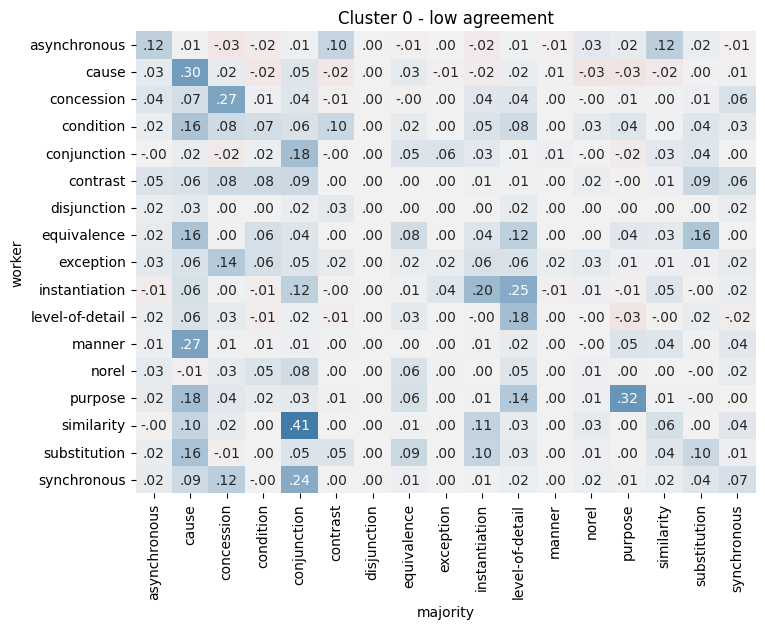}
    \includegraphics[width=0.9\linewidth]{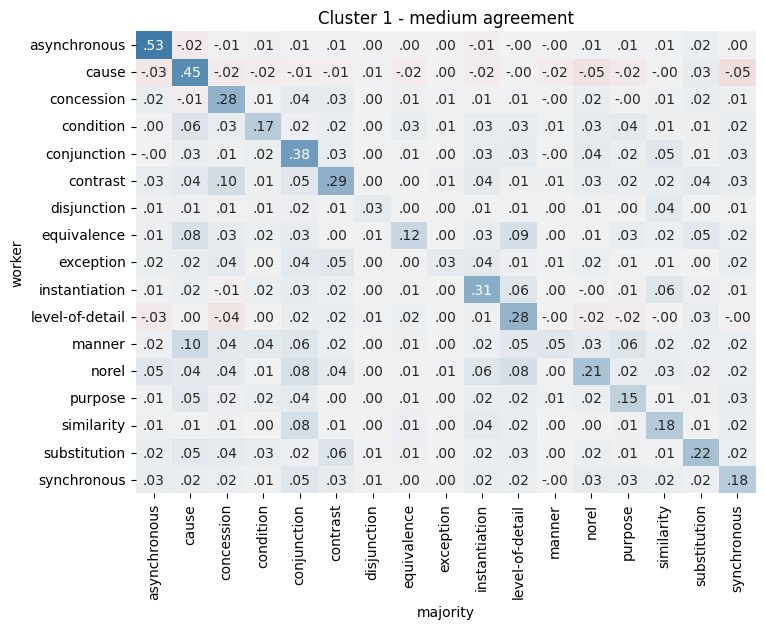}
    \includegraphics[width=0.9\linewidth]{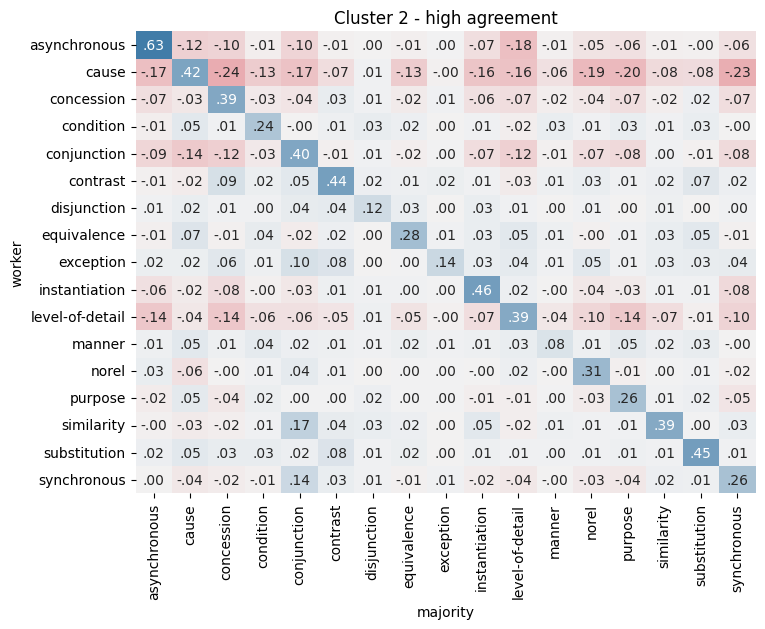}
    \caption{\small Mean nPMI values between the worker and majority labels for the three clusters of workers with different levels of agreement with the majority: Cluster 0 (low agreement), Cluster 1 (medium agreement), Cluster 2 (high agreement). }
    \label{fig:cluster}
\end{figure}

\begin{figure*}[htpb]
    \centering \small
    \includegraphics[width=0.245\linewidth]{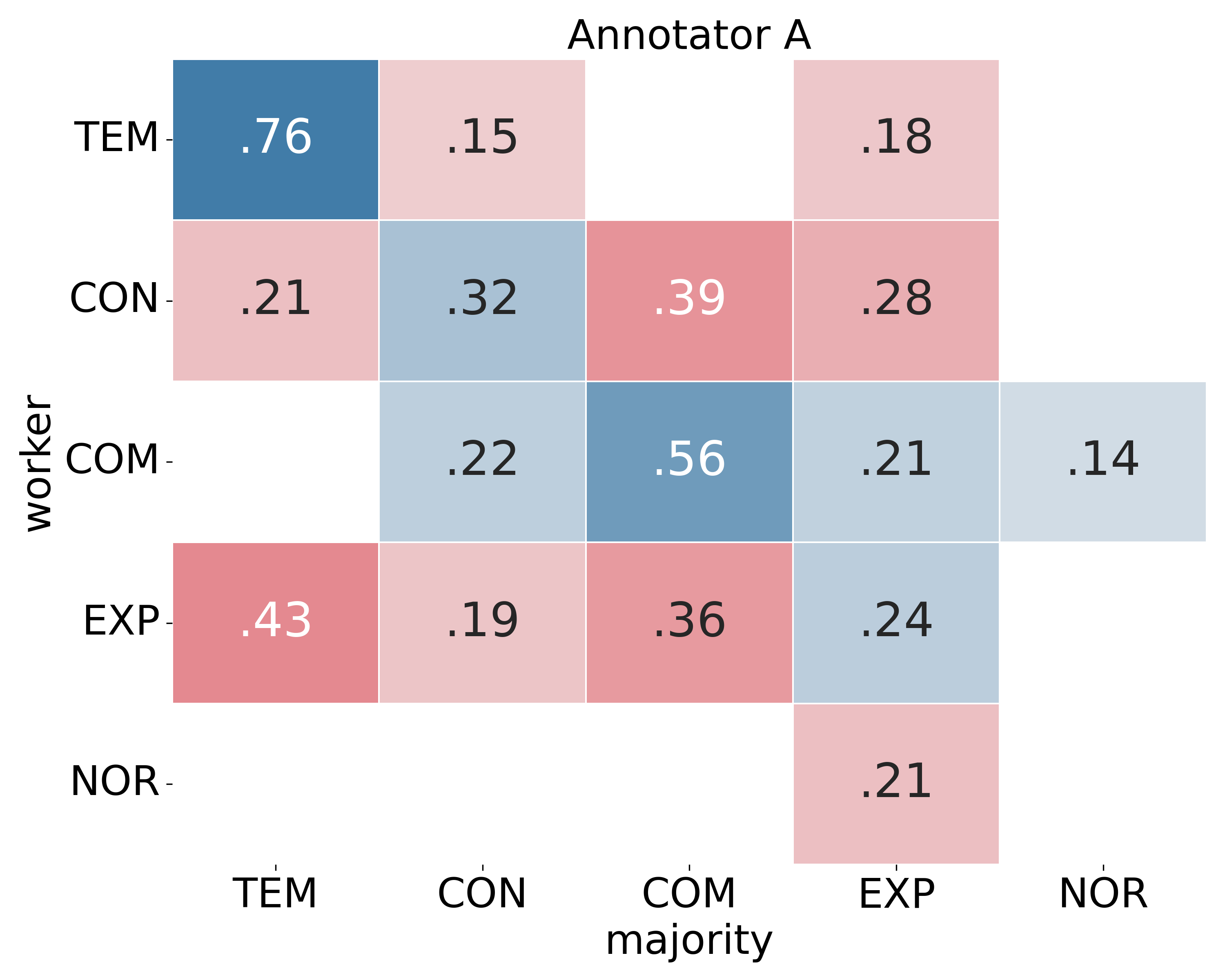}
     \includegraphics[width=0.245\linewidth]{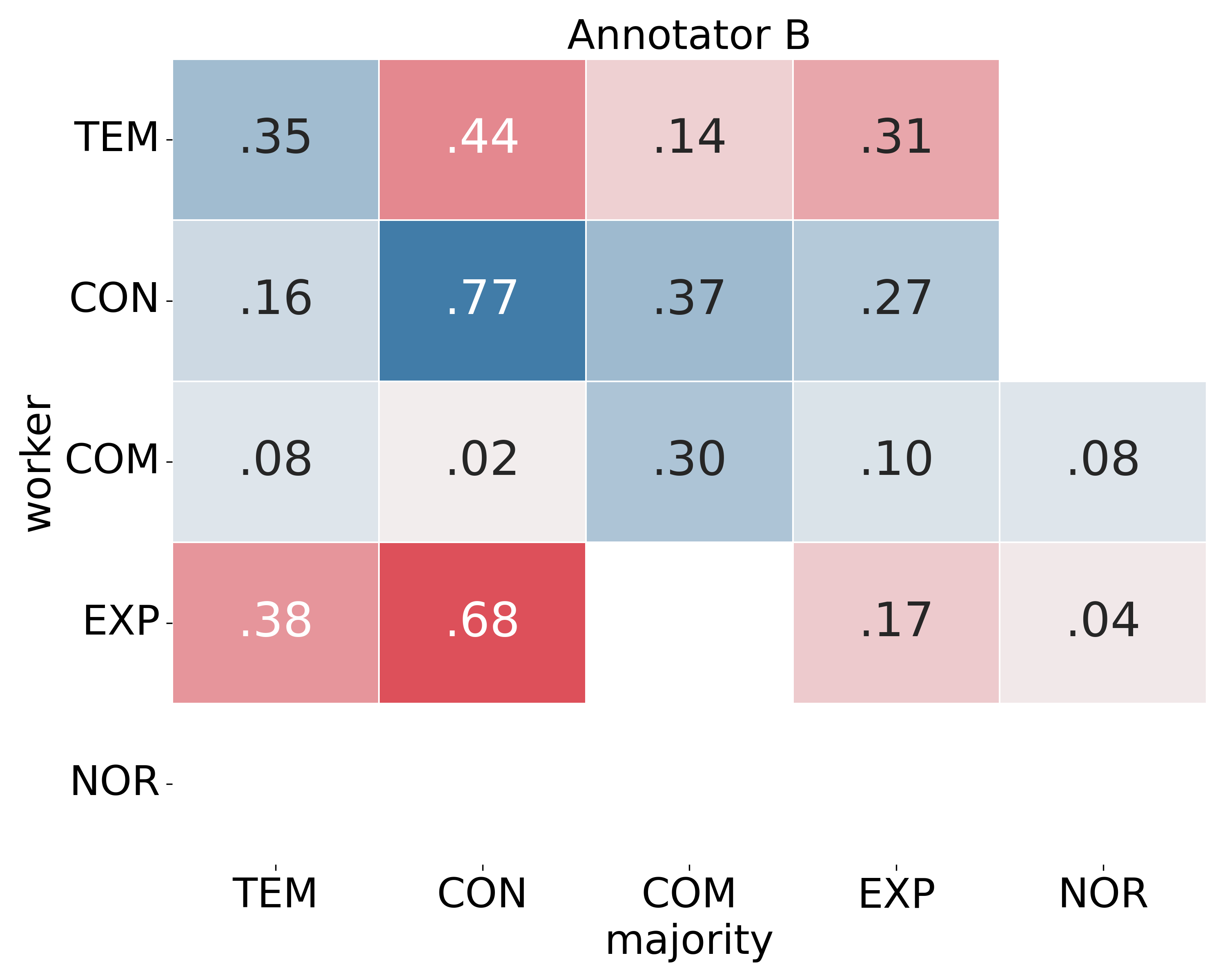}
     \includegraphics[width=0.245\linewidth]{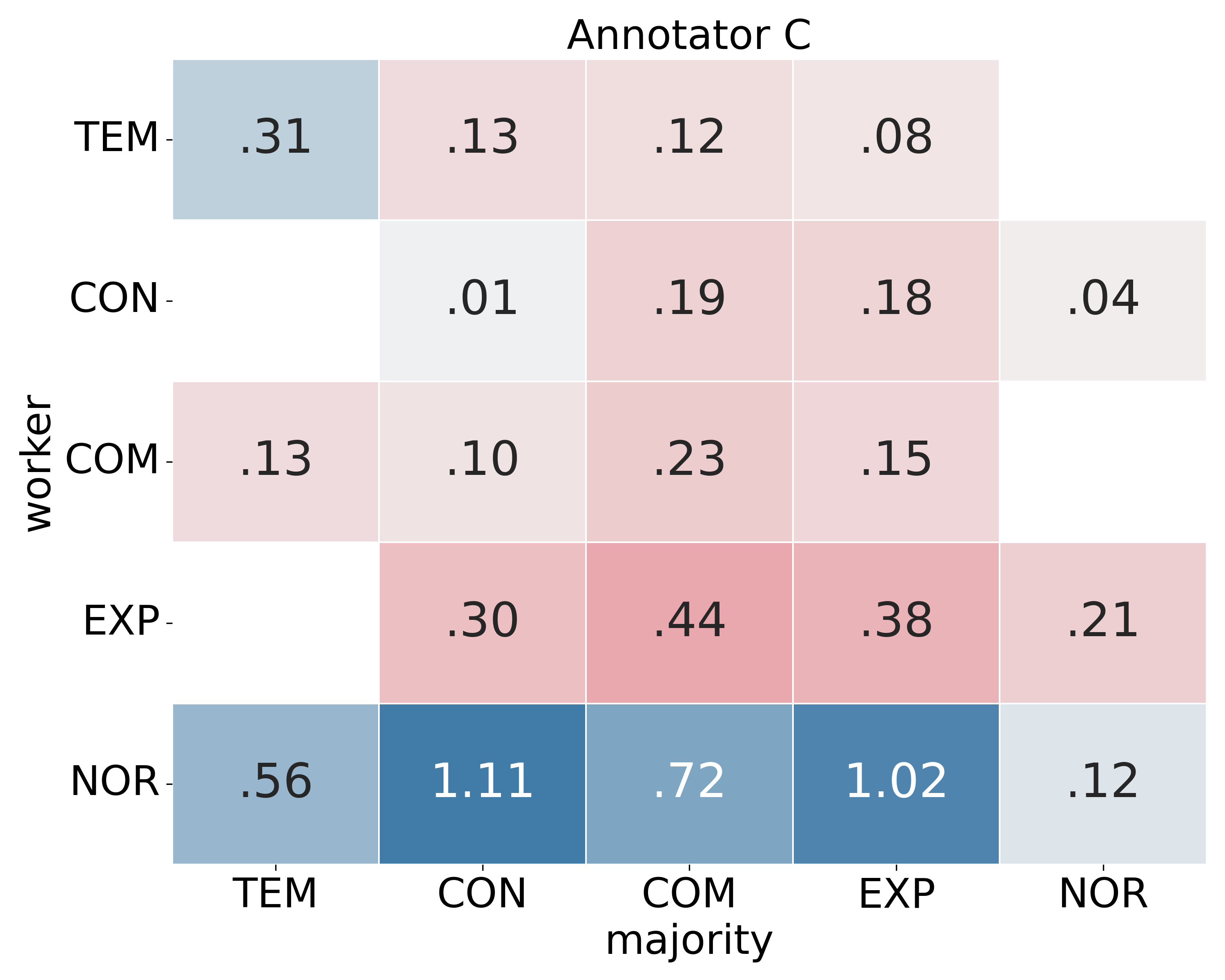}
     \includegraphics[width=0.245\linewidth]{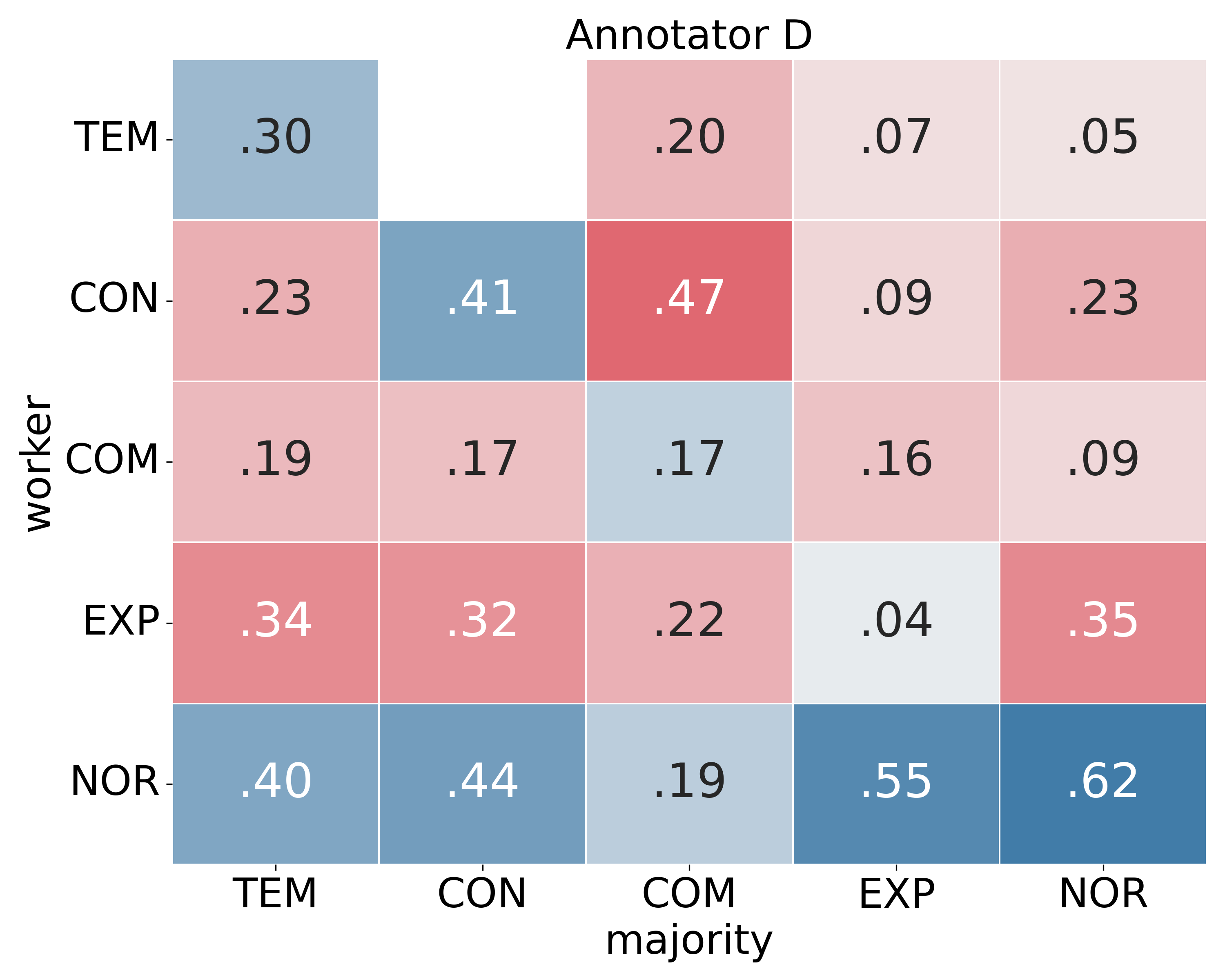}
    \caption{\small Bias of four workers compared with the majority label based on nPMI. \textbf{darker} colors mean more divergence from the majority, where \textbf{blue} means higher tendency (positive nPMI) and \textbf{red} means lower tendency (negative nPMI). Abbreviations: temporal, contingency, comparison, expansion, no relation.}
    \label{fig:NPMI}
\end{figure*}
\begin{figure*}[htpb]
    \centering \small
    \includegraphics[width=0.24\linewidth]{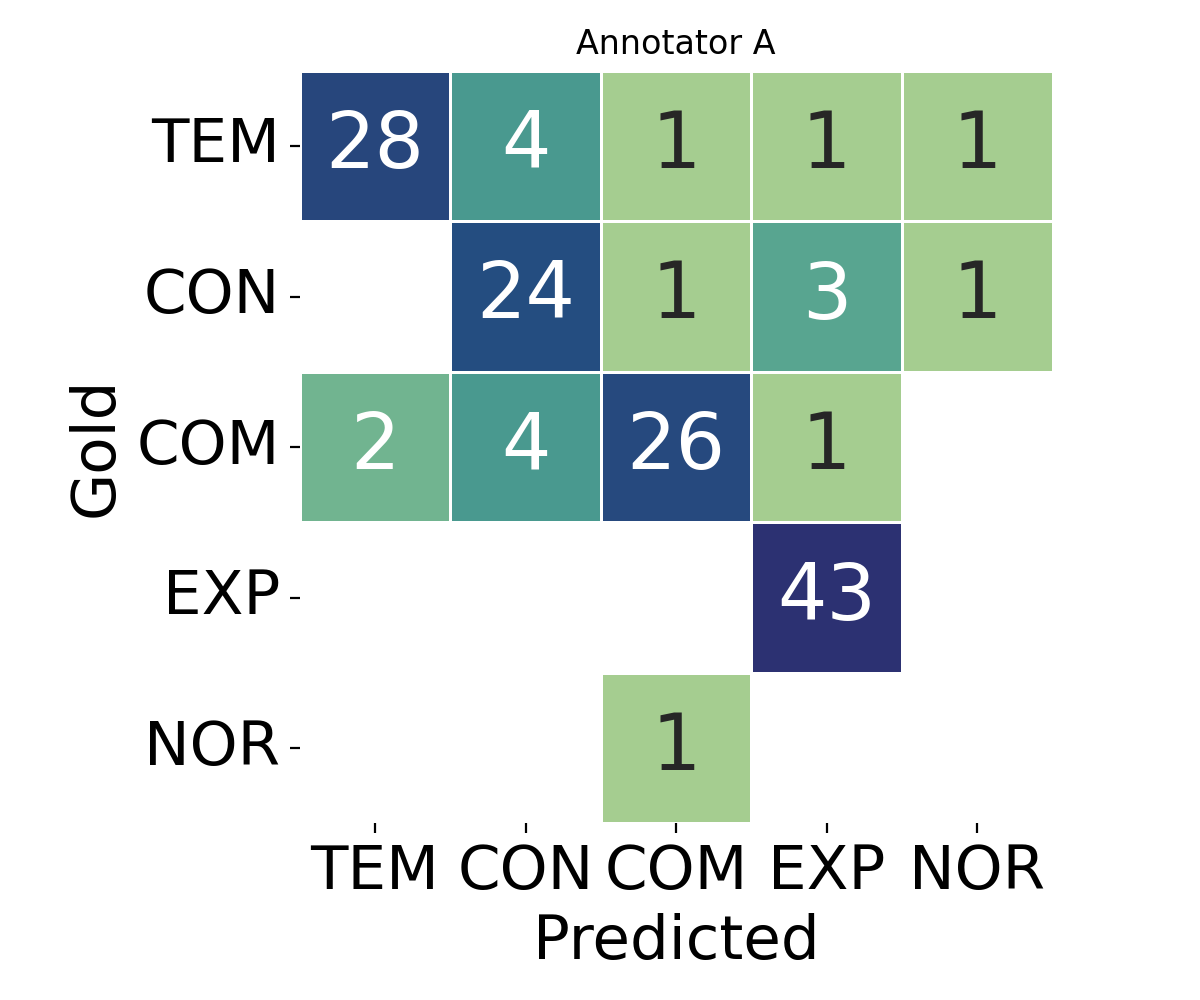}
     \includegraphics[width=0.24\linewidth]{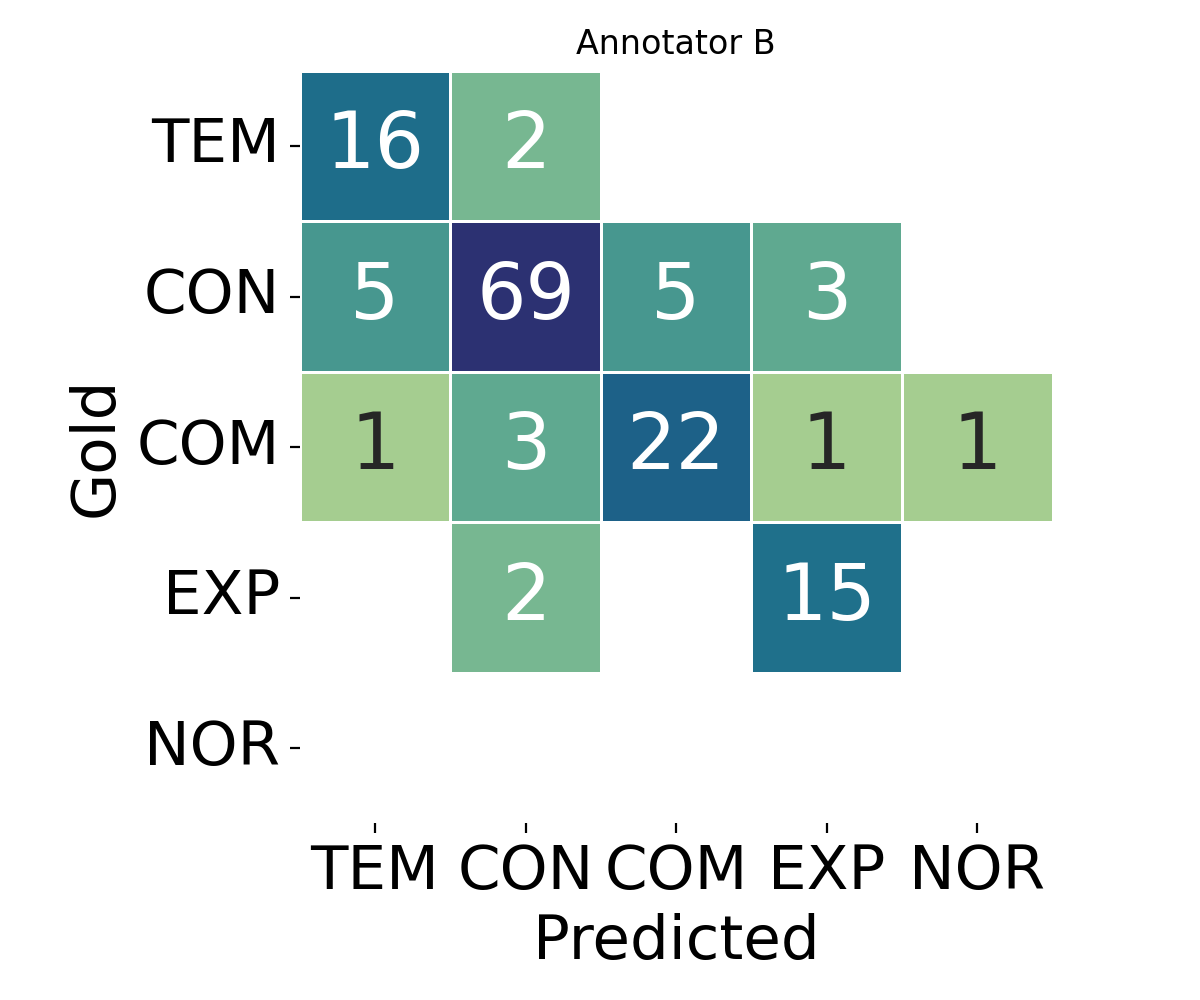}
     \includegraphics[width=0.24\linewidth]{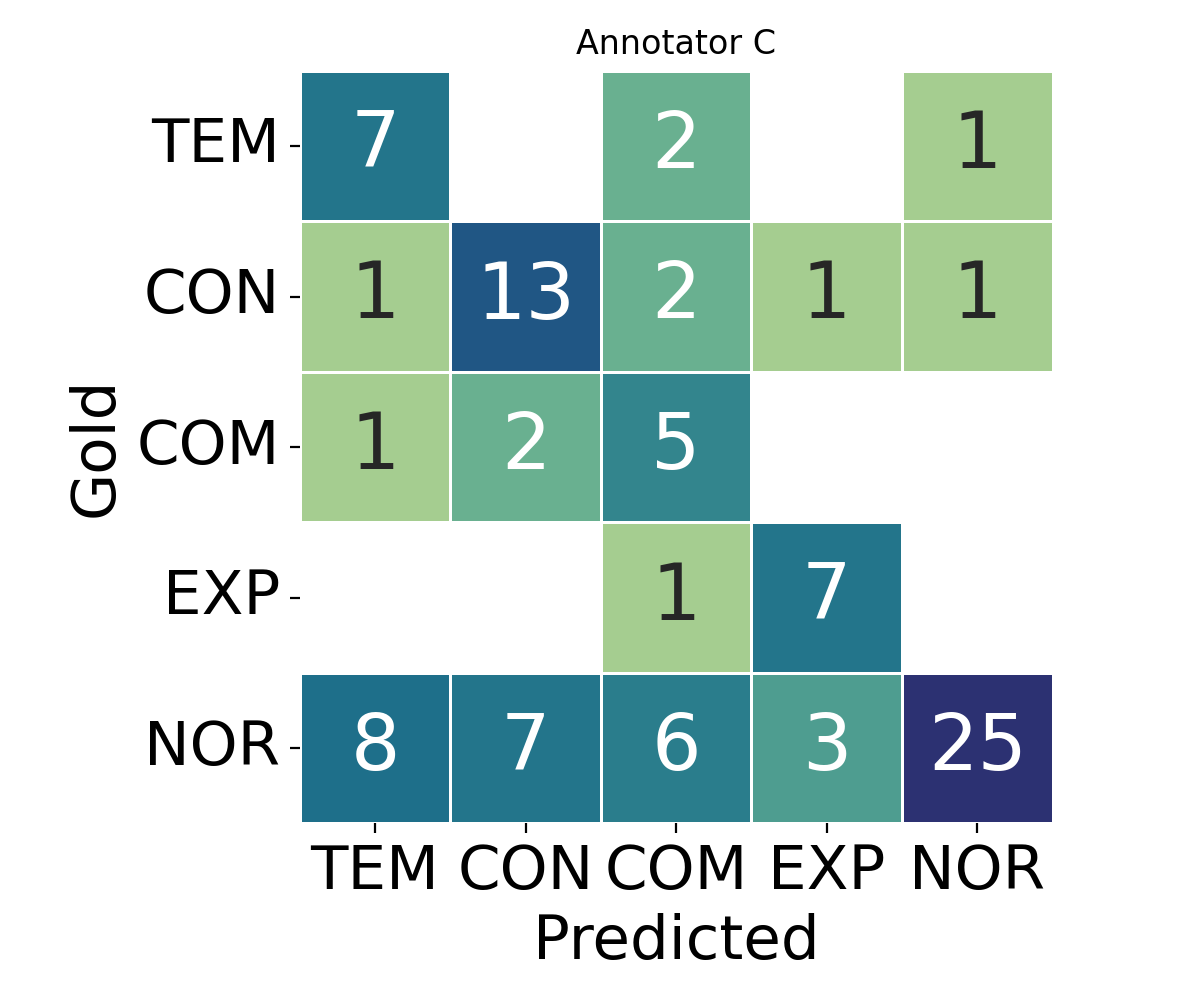}
      \includegraphics[width=0.24\linewidth]{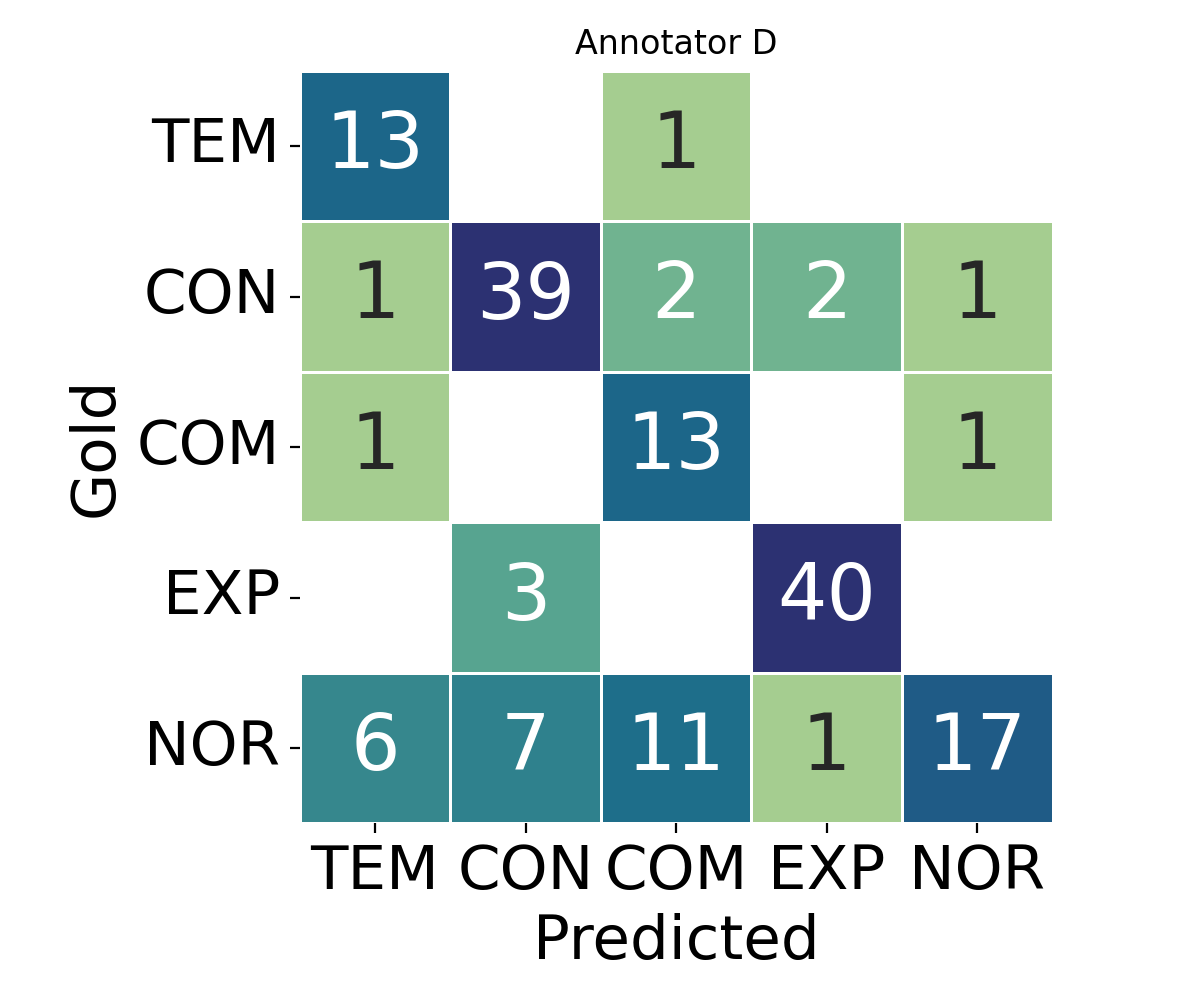}       
    \caption{\small Confusion matrices of worker-specific prediction of the above workers. \textbf{Gold} refers to the original labels. Abbreviations: temporal, contingency, comparison, expansion, no relation.}
    \label{fig:predict_cm}
\end{figure*}
Next, we compare model performance between workers with more conventional interpretations of DRs and those with less typical ones.
We capture this difference by comparing the worker's annotation with the majority. Specifically, we compute the normalized Pointwise Mutual Information (nPMI) between each of the $17$ Level-2 labels assigned by the worker ($L_{w}$) and the majority ($L_{maj}$) in the same subset of samples. A positive nPMI($L_{w}$,$L_{maj}$) indicates that a worker tends to assign $L_{w}$ to samples for which the majority label is $L_{maj}$, suggesting an association between this pair of worker and majority labels. Conversely, a negative nPMI value suggests that the worker tends \textit{not} to assign $L_{w}$ when the majority label is $L_{maj}$, reflecting a negative corelation between this label pair.

We then apply K-means clustering ($k=3$) to group workers based on their vectors of nPMI values. Figure~\ref{fig:cluster} illustrates the mean nPMI matrices for the three clusters. As shown, the workers in Cluster 0 show low agreement, frequently assigning labels that differ from the majority. In contrast, Cluster 2 workers exhibit the highest agreement with the majority vote, characterized by a dark blue main diagonal (indicating strong positive correlations) and light or red off-diagonal cells (indicating weak or negative correlations). Cluster 1 also demonstrates relatively high agreement, although the absence of strong red cells indicates occasional deviations from the majority consensus.  

The lower half of Table~\ref{tab:ablation} presents the performance of the AE model across these clusters of workers with different levels of agreement with the majority (AE$_{Low-\kappa}$, AE$_{Med.-\kappa}$, and AE$_{High-\kappa}$). The results reveal that the model performs notably worse in predicting the annotations of workers who have low agreement with the majority, implying that the labeling biases and perspectives of this group are harder for the model to learn and predict.
We further examine specific cases to understand when the model can effectively capture annotator preferences and when it fails to do so.
Figure~\ref{fig:NPMI} presents the nPMI matrices for $4$ workers, comparing their labels with the majority labels. Figure~\ref{fig:predict_cm} shows the corresponding confusion matrices of the predictions of the AE model against the actual labels of these workers. For clearer visualization, all values are aggregated at the Level-1 category.

Annotator~A exhibits a clear preference for the \textit{COMparison} relation, whereas Annotator B shows a bias toward \textit{CONtingency}, as indicated by the prominent blue rows in their respective nPMI matrices (Figure~\ref{fig:NPMI}). Importantly, the AE model successfully captures these tendencies: it correctly predicts when these workers are likely to choose these relations and when they are not, as reflected in the darker main diagonal indicating alignment between the predicted and actual labels in Figure~\ref{fig:predict_cm}.

In contrast, Annotators C and D display an apparent bias toward the \textit{NO-Relation label}, yet the AE model struggles to model this behavior accurately. Instead, it tends to under-predict the \textit{NO-Relation} label; the model predicts other classes while the workers actually had selected \textit{NO-Relation}. Manual inspection of these cases reveals that Annotators C and D often resort to the \textit{NO-Relation} label when the DR is particularly ambiguous or difficult to interpret (examples in Section~\ref{sec:qualitative}). This suggests that their use of \textit{NO-Relation} is not a consistent labeling bias but rather a fallback strategy employed in response to uncertainty. 

The analysis demonstrates that modeling individual perspectives in IDRR is particularly challenging when the workers demonstrate inconsistent behaviors in uncertain, difficult cases. To gain deeper insight into how such cases might be mitigated, we conduct a manual analysis of the frequency and nature of different types of disagreement in IDRR.
\subsection{Causes of disagreement}
\label{sec:qualitative}

We selected a subset of $100$ samples that were consistently annotated by $3$ workers, referred to here as Annotators X, Y, and Z. 
For each Level-2 label assigned by the workers, we examined the nature of the disagreement and categorized it according to the potential underlying reason for divergence. When a worker’s label matched one or more of the others, it was counted as a case of agreement. Table~\ref{tab:reasons} presents the distribution of disagreement categories, while Table~\ref{tab:top3} summarizes the three most frequently assigned labels by each worker within this subset.  We discuss these categories and their implications in detail below.

\begin{table}[ht]
    \centering \small
    \begin{tabular}{lll|l}
        \hline
         \multicolumn{3}{l|}{agreement} & $35\%$  \\
         \hline
         \multicolumn{3}{l|}{disagreement} &\\
         &\multicolumn{2}{l|}{inherent ambiguity} & $31\%$\\
         &\multicolumn{2}{l|}{cognitive demands:} &\\
         && unidentified cues & $5\%$\\
         && sentence complexity & $10\%$\\
         && annotation error & $11\%$\\
         &\multicolumn{2}{l|}{task design} & $8\%$\\
         \hline      
    \end{tabular}
    \caption{Distribution of disagreement categories in $300$ annotations ($100$ items $\times$ $3$ workers)}
    \label{tab:reasons}
\end{table}

%
\begin{table}[ht]
    \centering \small
    \begin{tabular}{@{}l|lll@{}}
    \hline
        & top1 & top2 & top3 \\
        \hline
        X & conj.($20\%$) & asynch.($13\%$)  & inst. ($10\%$) \\
        Y & lev-of-det ($22\%$)& cond. ($11\%$)  & synch.  ($9\%$) \\
        Z & cause ($23\%$) & lev-of-det ($11\%$) & goal ($9\%$) \\
        \hline
    \end{tabular}
    \caption{Top $3$ annotated labels by Annotators X, Y and Z in the $100$ analyzed items. Abbreviations: conjunction, asychronous, instantiation, condition, sychronous.}
    \label{tab:top3}
\end{table}
Table~\ref{tab:reasons} shows that over one-third of the annotations fall under \textbf{agreement}, where the assigned label matches at least one of the other workers for the same item. Another one-third of the annotations were found to differ due to the \textbf{inherent ambiguity} of the DRs. This ambiguity arises from a variety of causes.
A primary source of ambiguity stems from differences in how workers interpret or emphasize \textbf{different spans} within the relation arguments.
In Example (2), Annotator Z focused only on the beginning of Arg2, interpreting the relation as \textit{concession} (i.e., the man opened the window even though it was raining). In contrast, Annotator Y considered the full span of Arg2 and identified the DR as \textit{synchrony}, emphasizing the simultaneous nature of the two events.

\ex {Arg1: \textit{What was the man about to do? He lowered the window on our side.}
Arg2: A heavy rain was now falling, and, by a gesture, the man expressed his annoyance at his not having an umbrella...}

Another source of disagreement arises from the \textbf{lack of contextual information}. Because DiscoGeM is designed to capture the organic, local interpretation of DRs, 
workers are instructed to label relations given only one to two sentences of context preceding Arg1 and following Arg2. Consequently, the interpretation of a DR can vary depending on each worker’s inferred or assumed broader context.
In the narrative fragment of Example (3), the relationship between the characters and the nature of the events remain unclear without access to the wider context of the novel, leading to annotator disagreement. 

\ex {Arg1: \textit{Pianelli was constantly looking for a scoop, but I’d often wondered what lay behind his fascination with this particular case.}
Arg2: From what I remembered, he and Vinca had never been close. They’d never spent time together...}

%
%
%
In addition, workers also exhibit \textbf{individual preferences} when labeling ambiguous DRs.  In Example (4), both Arg1 and Arg2 refer to the same entity pair (Fanny and “I”), yet the DR between them is not obvious. This type of relation corresponds to what the PDTB framework classifies as an \textit{entity relation} \cite{prasad2017penn}, which is typically implicit in natural text and often resists representation by an explicit connective. In this case, Annotator X labeled the relation as \textit{conjunction}, whereas Annotator Y chose \textit{level-of-detail}. Both interpretations are reasonable and highlight distinct annotation preferences toward particular relation types, preferences that are also reflected in the individual label distributions shown in Table~\ref{tab:top3}.
\ex {Arg1: \textit{...Fanny was the first girl I ’d ever dated, back when I was fourteen and in the ninth grade.}
Arg2: We went to see a Saturday matinee of Rain Man...}

The types of ambiguity discussed above are also commonly observed in other annotation tasks, often leading to similar patterns of disagreement among annotators \cite{jiang-marneffe-2022-investigating,sandri-etal-2023-dont}. 
We additionally identified types of disagreement that emerge from the \textbf{cognitively demanding} nature of IDRR. 
First, some workers appear to be more sensitive to or linguistically skilled in generating discourse inferences based on subtle \textbf{cues}, while others may overlook these cues or rely less on them in their interpretations \citep[][]{graesser1994constructing,scholman2020individual,zwaan1998situation}. In Example (5), the specification of a year in the text prompted Annotators X and Y to select an \textit{asynchronous} relation. In contrast, Annotator Z focused on the event development, labeling the relation as \textit{cause}, which is also justifiable given the context.
\ex {Arg1: \textit{
...I lived in fear and in a constant state of mental anguish that led me to disastrously fail my final exams.}
Arg2: By the summer of '93, I had already left the Côte d’ Azur for Paris, where ... I enrolled in a second-rate business school.}
Another cognitive challenge in IDRR arises from the difficulty of interpreting \textbf{complex sentence} structures, which obscure the intended semantics of the discourse arguments. In Example (6), Arg2 begins with the connective \textit{after}, which actually signals an intra-argument temporal relation within Arg2 itself. However, this connective appears to have prompted Annotator X to label the relation between Arg1 and Arg2 as \textit{asynchronous}. The DR should instead concern the contrast between Maxime and "I", as identified by Annotator~Y.
\ex {Arg1: \textit{...While it had devastated me and stopped me in my tracks, for Maxime, it had broken down barriers, released him from a straitjacket, and left him free to write his own story.}
Arg2: After what happened, I was never the same...}
%
%
%
%
%
In our manual analysis, we also identified a portion of labels that we could not justify and therefore categorized as \textbf{annotation errors}. In Example (7), Annotator Y labeled the relation as \textit{equivalence},
which is not reasonable given the context. It is possible that they were uncertain about the correct relation and selected a label at random. Here, Annotator X assigned the label \textit{no-relation}, which is appropriate in this case. However, since this worker tends to select \textit{no-relation} in a range of cases, such as Example~1, it can also be argued that they may be using this label as a fallback strategy when unsure of the correct relation.
\ex {Arg1: \textit{From what I remembered, he and Vinca had never been close...
they had almost nothing in common.} Arg2: Vinca's mother was Pauline Lambert, an Antibes-born actress with close-cropped red hair...}

Finally, we observe that some of the disagreements in DiscoGeM may have arisen from the \textbf{task design}, specifically, from the choice of connectives to represent each DR type. In example (8),  \textit{condition} is a reasonable interpretation (i.e., If the report is on the table, we are determined to adopt it). However, the connective associated with this relation, \textit{in that case}, does not fit naturally in the context, discouraging workers from choosing it. In other words, the annotation outcome depends on the workers’ flexibility in performing the connective insertion task; some may prioritize selecting a connective that sounds more natural in the given context, even if it does not perfectly represent the underlying relation. 
\ex {Arg1: \textit{...The report has nevertheless been passed by the Committee for Foreign Affairs, it is on the table and it is essential to vote on the report... 
} Arg2: Our group is fully determined to debate and adopt it ...}

Overall, much of the observed variation stems from the cognitively demanding nature of IDRR, which leads to substantial disagreement that is difficult to model using perspectivist approaches due to inconsistencies in annotator behavior. 
The nature of shallow discourse parsing, where DRs are determined at the level of sentence pairs, allows room for individual differences in interpretation.

Nonetheless, these divergent perspectives are accounted for by existing theoretical principles. For instance, clause-level discourse can support multiple attachments to different anaphoric clauses \cite{webber-etal-2003-anaphora}, or interact with presuppositional structure \cite{asher2003logics}. In Example (2), the interpretation depends on whether ``was falling" attaches to ``lowered" or ``expressed." Similarly, in Example (4), “went to see (a movie)” may be interpreted either a ``date" or as a separate event.

When aggregated to the coarser Level-1 relation classes, there remains a higher likelihood that noisy labels overlap with consciously assigned labels. However, at the more fine-grained Level-2 classes, the noise appears to amplify disagreement, making individual perspectives harder to model consistently and resulting in markedly lower performance.
\section{Conclusion}

IDRR is a task where disagreement arises not only from the inherent ambiguity of language and differences in background knowledge but also from individual variation in the cognitive ability to interpret DRs. In this work, we evaluated existing approaches to model disagreement in IDRR with varying levels of precision. We found that models designed to predict specific annotator perspectives experience substantial performance drops when the prediction classes are fine-grained. In such cases (i.e., when the number of classes is large), modeling general per-instance label distributions offers a more realistic and reliable estimation.

Further analysis reveals that annotators behave inconsistently when faced with  cognitively demanding cases; e.g., they may use the \textit{no-relation} tag when unable to identify a suitable connective. To reduce such noise, annotators could be given the option to skip uncertain cases during data collection; however, this approach is often difficult to implement reliably in a crowd-sourcing setting, as it may be prone to misuse. 
Another possible direction is to distinguish genuine interpretive variation from annotation errors \cite{10.1162/coli_a_00464,weber2024-varierrnli}; particularly, annotation models may prove of use \cite{hovy-etal-2013-learning,passonneau2014benefits,ivey-etal-2025-nutmeg}. However, our findings suggest that in IDRR, the inherent ambiguity may blur the boundary between the two as well. In future work, we plan to explore methods that integrate annotation error detection with perspectivist modeling to more effectively capture human label variation in highly ambiguous discourse interpretation.

\section{Acknowledgments}
This project is supported by the German Research
Foundation (DFG) under Grant SFB 1102 (``Information Density and Linguistic Encoding", Project-
ID 232722074); and by NWO under the AINed Fellowship Grant NGF.1607.22.002  (``Dealing with Meaning Variation in NLP").

\section{Limitations}
All of the modes compared in our paper are standard architectures using RoBERTa-base \cite{liu2019roberta} as a backbone. 
They may not reflect the current state of the art in neural approaches to IDRR. We pick RoBERTa as a generic encoder-type model for the sake of demonstration. The same results may not hold using other advanced IDRR models.

The subset of data that we used for quantitative analysis may be insufficient to cover all the peculiarities of individual annotations. However, given that all DiscoGeM annotators worked on randomly batched items, parallel data is sparse, and we sampled the largest existing subset thereof. We consider it representative of the original perspectives, as it covers a diverse set of labels and elicits meaningful insights into the annotators' decision making. We leave further analysis into reasons for disagreement to future work.

\section{Bibliographical References}
\bibliographystyle{lrec2026-natbib}
\bibliography{lrec2026}

\section{Language Resource References}
\bibliographystylelanguageresource{lrec2026-natbib}
\bibliographylanguageresource{languageresource}

\end{document}